\documentclass{article}

\PassOptionsToPackage{numbers, compress}{natbib}

\usepackage[preprint]{neurips_2025}

\usepackage[utf8]{inputenc} %
\usepackage[T1]{fontenc}    %
\usepackage{hyperref}       %
\usepackage{url}            %
\usepackage{booktabs}       %
\usepackage{amsfonts}       %
\usepackage{nicefrac}       %
\usepackage{microtype}      %
\usepackage{xcolor}         %
\usepackage{subcaption}  %
\usepackage{graphicx}    %
\usepackage{makecell}
\usepackage{subcaption}   %
\usepackage{caption}

\newcommand{\centercell}[1]{\begin{tabular}[c]{@{}c@{}}#1\end{tabular}}

\usepackage{amsmath, amssymb}

\usepackage{graphicx}

\usepackage{xcolor}  %

\newcommand{\iliasScale}{0.71}

\newcommand{\n}{QuARI}

\title{QuARI: Query Adaptive Retrieval Improvement}

\author{Eric Xing$^1$ \qquad Abby Stylianou$^2$ \qquad Robert Pless$^3$ \qquad Nathan Jacobs$^1$\\
$^1$Washington University in St.\ Louis\quad $^2$Saint Louis University\quad $^3$The George Washington University}

\begin{document}

\maketitle
\begin{abstract}

Massive-scale pretraining has made vision-language models increasingly popular for image-to-image and text-to-image retrieval across a broad collection of domains. However, these models do not perform well when used for challenging retrieval tasks, such as instance retrieval in very large-scale image collections. Recent work has shown that linear transformations of VLM features trained for instance retrieval can improve performance by emphasizing subspaces that relate to the domain of interest. In this paper, we explore a more extreme version of this specialization by learning to map a given query to a query-specific feature space transformation.  Because this transformation is linear, it can be applied with minimal computational cost to millions of image embeddings, making it effective for large-scale retrieval or re-ranking.  Results show that this method consistently outperforms state-of-the-art alternatives, including those that require many orders of magnitude more computation at query time.

\end{abstract}

\section{Introduction}

Recent advances in language-image pretraining have significantly improved performance in various vision-language tasks, including image-to-image and text-to-image retrieval. These models learn to align images and their corresponding textual descriptions in a shared embedding space over large-scale datasets. Yet, despite their success, they show notable limitations in retrieval scenarios.

Pretrained models often rely on global image features that encapsulate the overall content of an image. While effective for general classification tasks, these global representations may not capture fine-grained details essential for distinguishing between visually similar images based on specific textual queries~\citep{zhong2021regionclip}. This limits retrieval performance when nuanced differences are critical.

To address the shortcomings of global feature representations, re-ranking methods have been proposed, which involve a secondary, more expensive analysis of the top retrieval candidates. Techniques such as two-stage retrieval pipelines~\citep{pinecone2023rerankers} and re-ranking transformers~\citep{tan2021reranking} aim to refine initial retrieval results. However, these approaches often entail substantial computational overhead, making them impractical for real-time applications or large-scale deployments.

\begin{figure}[t!]
    \centering
    \includegraphics[width=\linewidth]{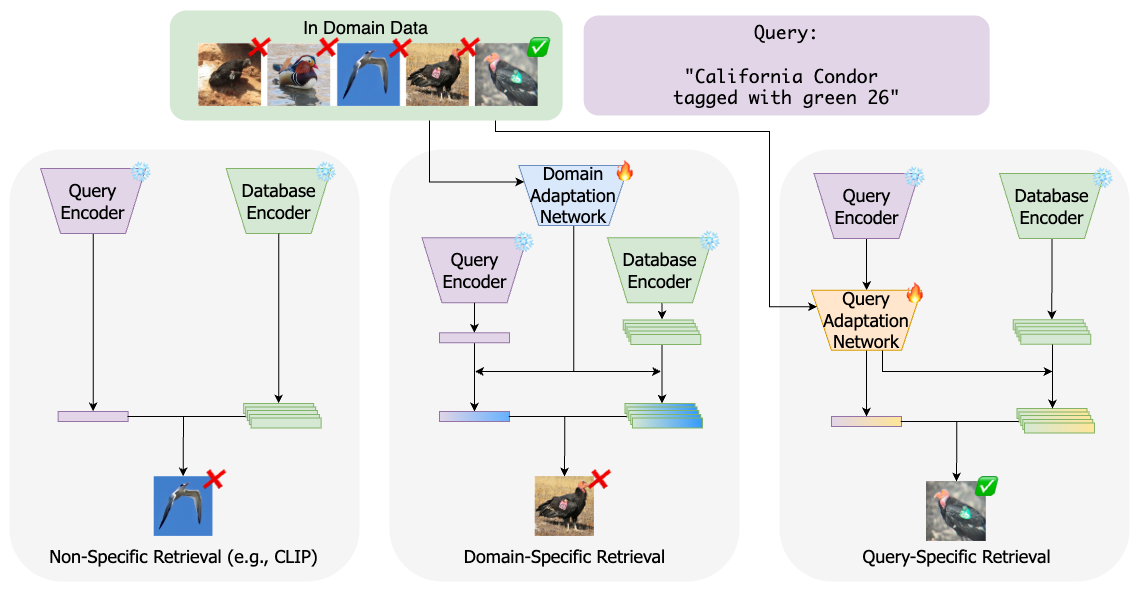}
    \caption{We propose a new query-specific approach to retrieval, {\n}. {\n} dynamically adapts embeddings \emph{per-query} to significantly improve retrieval performance compared to non-specific retrieval with general purpose embedding features like CLIP, and domain-specific retrieval with transformations learned for a specific domain, with little computational overhead. Figure~\ref{fig:arch} shows the details of the Query Adaptation Network.}\vspace{-.75em}
    \label{fig:teaser}
\end{figure}

In this work, we propose a novel approach that integrates \textit{query-specific} embedding projections into the retrieval process. By dynamically adjusting the embedding space based on the input, our method prioritizes the most relevant fine-grained semantic alignments between text and image features for each specific query. Figure~\ref{fig:teaser} contrasts our query-specific retrieval approach with non-specific retrieval using a foundational vision-language model like CLIP~\citep{radford2021learning} and domain-specific retrieval that has been adapted for a given dataset or task. Our framework,  Query Adaptive Retrieval Improvement ({\n}), enhances retrieval accuracy without incurring the high computational costs associated with traditional re-ranking methods. Our specific contributions are as follows:
\begin{itemize}
    \item We identify and articulate the limitations of contrastively pretrained models in capturing fine-grained details necessary for accurate retrieval, and analyze the inefficiencies of existing re-ranking methods and their impact on retrieval performance.
    \item We introduce {\n}, an embedding projection framework that adapts global representations to individual queries, improving accuracy while maintaining computational efficiency.
    \item We demonstrate that {\n} yields large improvements in retrieval accuracy on multiple extremely challenging retrieval tasks.
\end{itemize}

\section{Related Work}

\paragraph{Text-Image Pretraining and Global Embeddings. }Foundation models based on language-image pretraining, such as CLIP~\citep{radford2021learning} and the SigLIP family of models ~\cite{tschannen2025siglip2multilingualvisionlanguage,Zhai_2023_ICCV_ogsiglip}, have achieved remarkable success in aligning visual and textual modalities. These models learn global representations by maximizing the similarity between paired image-text inputs. However, the reliance on global features can be detrimental in retrieval tasks that require fine-grained discrimination. Methods like RegionCLIP~\citep{zhong2021regionclip} attempt to address this by incorporating region-level features, but challenges remain in capturing nuanced details essential for accurate retrieval.
FILIP introduces token‑wise late interaction to capture patch–word similarities~\citep{yao2021filip}, while SPARC sparsifies such interactions for efficiency in large‑scale pre‑training~\citep{sparc2024}. Region‑centric pre‑training further bridges image‑level and region‑level semantics for open‑vocabulary detection~\citep{kim2024regioncentric}.

These methods highlight a growing recognition of the limitations of global embeddings and attempt to address them through increasingly sophisticated alignment mechanisms. However, they still largely rely on a single static representation per image. In contrast, our method is dynamic, modifying the representation space per query to prioritize the learned aspects that are relevant for the query while de-emphasizing irrelevant aspects.

\paragraph{Efficient Late‑Interaction and Re‑ranking Paradigms. }Late‑interaction retrievers (e.g., ColBERT~\citep{khattab2020colbert}) decouple encoder computation from expensive cross‑attentions, enabling scalable yet fine‑grained passage search.  In vision, there have been a variety of approaches to re-ranking, including local feature based geometric re-ranking approaches~\citep{philbin2007object,shao2023global,sharif2014cnn,Suma_2024_ECCV_ames}, and the recent Re-ranking Transformers~\citep{tan2021reranking}, which refines top‑$k$ candidates with lightweight self‑attention. %

Methods like two-stage retrieval pipelines~\citep{pinecone2023rerankers} and re-ranking transformers~\citep{tan2021reranking} refine initial retrieval results by re-evaluating top candidates with more sophisticated models. While effective, these approaches introduce significant computational overhead, making them unsuitable for real-time applications or large-scale systems. Our approach instead shifts the complexity into a lightweight projection computed once per query. The approach is computationally lightweight enough that it can be used to re-rank very large numbers of candidates, or even entire datasets.

\paragraph{Hypernetworks. }Hypernetworks, or networks that predict the weights of other networks, were first introduced by Ha \emph{et al.} to generate recurrent neural network parameters on the fly~\citep{ha2016hypernetworks}. Recent advancements have explored the use of hypernetworks for personalization in generative and task-specific models. HyperDreamBooth~\citep{ruiz2023hyperdreambooth} introduces a hypernetwork that generates customized weights for text-to-image diffusion models, enabling the synthesis of subject-specific images with minimal data. HyperCLIP~\citep{hyperclip2023} employs a hypernetwork to generate the weights of a task-specific image encoder. These approaches demonstrate the potential of hypernetworks in capturing customized semantics efficiently. However, reliance on task-level customization of entire encoders, as in HyperCLIP, yields only modest performance gains on challenging retrieval tasks. Generating complete sets of encoder weights is both computationally expensive and difficult to optimize. In contrast, we show that using hypernetworks to adapt off-the-shelf features with lightweight, query-specific transformations can achieve strong performance without significant computational overhead.

\paragraph{Transformers as Hypernetworks. }The concept of using transformers as hypernetworks has been explored in various domains. For example, Transformers as Meta-Learners~\citep{yin2022transformers} leverage transformers to predict the weights of implicit neural representations, showcasing their capability in dynamic weight generation. 
We draw inspiration from these works, employing a transformer to predict query-specific projection matrices. In the following section, we discuss how we tailor this idea to the retrieval setting.

\section{Methodology}

Our approach introduces a \emph{transformer-based hypernetwork} for customized text-to-image retrieval. Unlike traditional methods relying on static embeddings, {\n} predicts customized linear projections conditioned on each query embedding to adapt database features for each query.

\subsection{Hypernetwork-Augmented Retrieval}
We begin by extracting global embeddings for queries and candidate images using a pretrained vision-language model, such as CLIP~\citep{radford2021learning} or SigLIP~\citep{Zhai_2023_ICCV_ogsiglip}. Formally, for a query $q$ and a set of ``gallery" images $\mathcal{G}=\{I_n|n=1, 2,...,N\}$, we obtain embeddings for a query $\mathbf{q}_i = \text{Enc}(q_i)$ and a database of gallery embeddings $\mathcal{D}=\{\mathbf{d}_1, \mathbf{d}_2, ..., \mathbf{d}_N\}=\{\text{Enc}(I_1), \text{Enc}(I_2), ... \text{Enc}(I_N)\}$. We learn a hypernetwork $H_\theta$ as follows:
\vspace{-.25em}\begin{equation}
    (\mathbf{q'}_i, T_i) = H_{\theta}(\mathbf{q}_i),
\end{equation}

Which outputs a transformed query representation $\mathbf{q'}_{i}$ and a transformation function $\mathbf{T}_{i}$. Retrieval is then performed by transforming the database:
\begin{align}
    \mathcal{D'}_i&
    =\{\mathbf{d'}_1, \mathbf{d'}_2, ..., \mathbf{d'}_N\} \\
    &=\{ T(\mathbf{d}_1), T(\mathbf{d}_2),  ..., T(\mathbf{d}_N) \}
\end{align}
The transformed database of embeddings $\mathcal{D'}_i$ may then be used to perform retrieval by selecting database indices maximizing some embedding similarity between the query $\text{sim}(\mathbf{q'}, \mathbf{d'}_i)$.

\begin{figure}
    \centering
    \includegraphics[width=\linewidth]{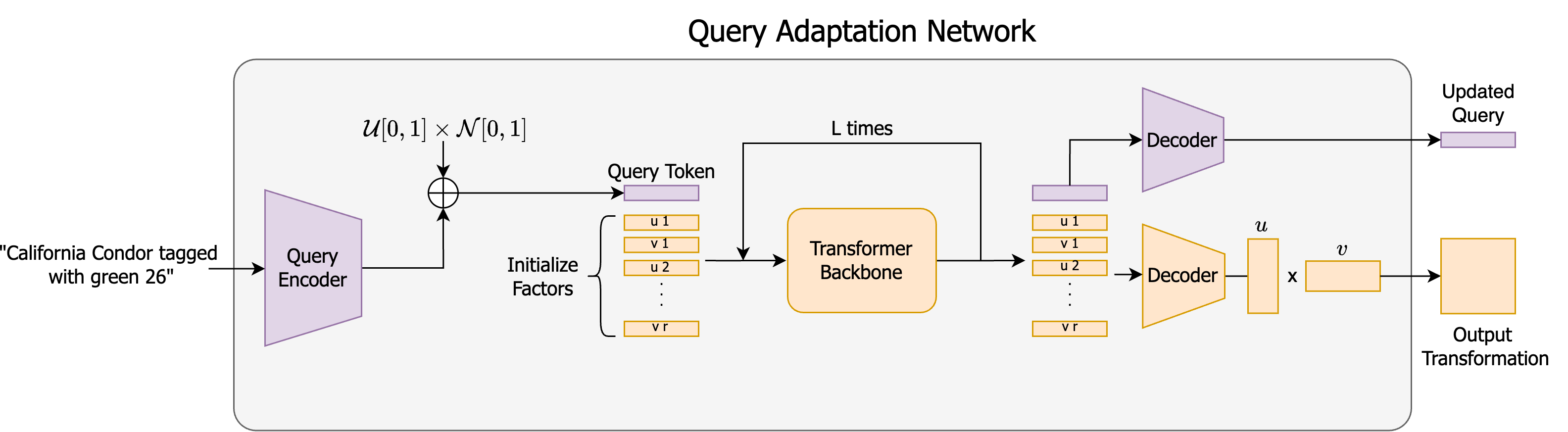}
    \caption{An overview of our query adaptation network. A zero initialization of the transformation matrix is tokenized by columns and passed to a transformer backbone with a conditioning token to obtain refined columns. This process is repeated $L$ times, refining the transformation. }\vspace{-.75em}
    \label{fig:arch}
\end{figure}

\subsection{Hypernetwork Architecture}

Our query adaptation network, shown in Figure~\ref{fig:arch}, is a hypernetwork which we will denote as $H_{\theta}$. It is a transformer-based module that conditions on the query embedding $\mathbf{q} \in \mathbb{R}^E$ and predicts both a customized query embedding $\mathbf{q}' \in \mathbb{R}^E$ and a transformation matrix $T \in \mathbb{R}^{E \times E}$ to adapt image embeddings of dimension $E$. To constrain computation and promote generalization, we parameterize $T$ as a low-rank matrix of rank $r=64$.

\paragraph{Tokenization of the Projection Matrix.} 
We define the transformation matrix $T$ via a learned low-rank decomposition:
\vspace{-.5em}\begin{equation}
    T = \sum_{j=1}^{r} \mathbf{u}_j \mathbf{v}_j^\top,
\end{equation}
where each pair $(\mathbf{u}_j, \mathbf{v}_j) \in \mathbb{R}^E \times \mathbb{R}^E$ represents a learned rank-one component of the projection matrix. These pairs are generated from a shared set of column-wise tokens, which are iteratively refined via a transformer encoder. Intuitively, each token pair $(\mathbf{u}_j, \mathbf{v}_j)$ defines a single direction in the transformed feature space, contributing one semantic feature to the customized representation.

\paragraph{Initialization and Conditioning.}
We initialize a bank of $2r$ tokens per sample in the batch: $r$ \emph{U-tokens} for projecting into the output space and $r$ \emph{V-tokens} for selecting features from the input space. The initial token representations are zero-initialized and refined over $L$ denoising steps. To condition the generation process on the query embedding, we first encode $\mathbf{q}$ using a two-layer MLP and add a learned timestep embedding to capture iterative update dynamics.

\paragraph{Iterative Refinement with Transformers.}
At each step $t \in \{1, ..., L\}$, we concatenate the query-conditioned control token with the current U/V token sequence and apply a shared transformer encoder. We also apply sinusoidal positional encodings~\cite{10.5555/3295222.3295349_attnisallyouneed} to the token sequence. 
The output of the transformer updates the tokens via residual addition:
\begin{align}
    \mathbf{u}_j^{(t+1)} &= \mathbf{u}_j^{(t)} + \Delta \mathbf{u}_j^{(t)}, \\
    \mathbf{v}_j^{(t+1)} &= \mathbf{v}_j^{(t)} + \Delta \mathbf{v}_j^{(t)}.
\end{align}

\paragraph{Decoding to Projection Matrices.}
Once the tokens are fully refined, we decode each U-token and V-token using separate MLP decoders to produce vectors in $\mathbb{R}^E$. The final transformation matrix is then constructed as:
\vspace{-.25em}\begin{equation}
    T = \sum_{j=1}^r \text{MLP}_\text{u}(\mathbf{u}_j) \cdot \text{MLP}_\text{v}(\mathbf{v}_j)^\top.
\end{equation}
The customized query representation $\mathbf{q'}$ is decoded from the control token by another MLP. 

\subsection{Training}
We train the model using a symmetric contrastive loss over the transformed query and image embeddings. Given a minibatch of $B$ text-image embeddings $\{(\mathbf{q}_i, \mathbf{d}_i)\}_{i=1}^B$, we first add noise to every query embedding to help bridge the text-image modality gap. We borrow this formulation from LinCIR~\citep{gu2024lincir}. This acts as a regularizer that also allows high performance with image queries, training only on text-image datasets. 
\begin{equation}
    \mathbf{q}_i \gets \mathbf{q}_i + \mathcal{U}[0, 1] \times \mathcal{N}[0, 1]
\end{equation}

Then, we compute the customized query text embedding $\mathbf{q'}_i$ and the personalization matrix $T_i$ using the hypernetwork $H_{\theta}$. 

\paragraph{Semi-Positive Sample Mining. }In the standard contrastive formulation, there is only one positive sample for each negative sample. This leads the learned transformations to overfit to the query. This results in samples that are completely different from the target and samples which share some, but not all, attributes with the target having near-equivalent similarities to the transformed query.

In order to address this, we also compute a set of ``semi-positive" samples $\mathcal{P}_i$ for each target image.  For every target image we compute a set of 100 nearest-neighbors using precomputed backbone embeddings and select the top 2 as ``semi-positives.'' We apply a softmax over the distribution of these 100 cosine similarities and use their logits as target similarity values. 

To train {\n}, we optimize a symmetric contrastive objective, as is standard for retrieval with semi-positive samples:

\begin{equation}
p_{i,j} = \frac{\exp(S_{i,j})}{\displaystyle\sum_{k=1}^{B} \exp(S_{i,k})},
\qquad
q_{j,i} = \frac{\exp(S_{j,i})}{\displaystyle\sum_{k=1}^{B} \exp(S_{k,i})},
\label{eq:softmax_probs} 
\qquad
S_{i,j} = \frac{\mathbf{q'}_i \cdot \mathbf{d}_j^i}{\tau},
\end{equation}

\begin{equation}
\mathcal{L} =
\frac{1}{2B} \sum_{i=1}^{B} \left[
- \sum_{j=1}^{B} \alpha_{i,j} \log p_{i,j}
- \sum_{j=1}^{B} \alpha_{j,i} \log q_{j,i}
\right],
\qquad
\hspace{-0.2in}\alpha_{i,j} =
\begin{cases}
1           & \text{if } j = i \text{ (positive)},\\
w_{i,j}     & \text{if } j \in \mathcal{P}_i \text{ (semi‑positive)},\\
0           & \text{otherwise},
\end{cases}
\label{eq:symmetric_contrastive}
\end{equation}

Where $\mathbf{d}_i^j$ is the $i$-th target image embedding transformed by the $j$-th personalization transformation $T_j$, and $\tau$ is a temperature parameter. We only consider similarities where the embedding similarity of a transformed query $\mathbf{q'}_i$ is only computed with target image embeddings transformed by $T_i$. 

The incorporation of semi-positive samples discourages the behavior of overfitting to training query-target pairs, and encourages lower ranked images that seem visually similar to the target to be returned with higher similarity than images that lack visual similarity.

\paragraph{Implementation Details}
We use the AdamW optimizer~\citep{loshchilov2017decoupled} with a cosine annealed learning rate cycling between \texttt{1e-5} and \texttt{2e-7} and a weight decay of \texttt{1e-2}. Experiments were conducted on a NVIDIA H100 with 80GB of VRAM. At inference, retrieval is performed by computing cosine similarity scores between the L2-normalized customized embeddings of a query and database images.

\section{Evaluation}

\paragraph{Evaluation Datasets. }We evaluate our framework on two challenging benchmarks: ILIAS and INQUIRE. ILIAS (Instance-Level Image retrieval At Scale) is a large-scale dataset designed to assess instance-level image retrieval capabilities~\cite{kordopatiszilos2025iliasinstancelevelimageretrieval}. It has 1,000 object instances, each represented by query and positive images, totaling 5,947 manually collected images. To evaluate retrieval performance under large-scale settings, ILIAS includes 100 million distractor images from YFCC100M~\cite{10.1145/2812802_yfcc}. It also includes a retrieval task over a curated set of 5M distractor images, and an image-to-image retrieval reranking task. 

INQUIRE~\cite{vendrow2024inquire} is a text-to-image retrieval benchmark tailored for expert-level ecological queries. It is built on the iNaturalist 2024 dataset, containing five million natural world images across 10,000 species. The benchmark features 250 expert-crafted queries spanning categories such as species identification, context, behavior, and appearance. INQUIRE evaluates two retrieval tasks: INQUIRE-Fullrank, requiring models to perform retrieval over the entire dataset, and INQUIRE-Rerank, focusing on refining initial retrieval results. 

\paragraph{Training Datasets. } Our framework can be trained on any paired text-image dataset. We utilize Microsoft Common Objects in Context (MS COCO)~\cite{lin2014microsoft}, Conceptual Captions 12M, and synthetically augmented BioTrove~\cite{yang2024biotrove} to train {\n}. MS COCO includes over 330,000 images annotated with five human-written captions per image. Conceptual Captions 12M~\cite{changpinyo2021conceptual} is a collection of approximately 12 million image-text pairs harvested from the web. BioTrove~\cite{yang2024biotrove} contains 161.9 million images across approximately 366,600 species, each annotated with taxonomic data. We extract a random subset of 5M samples from BioTrove for training. Since the BioTrove dataset only includes taxonomic and common-name level annotations, we use \texttt{Qwen2.5-VL-7B-Instruct} to caption a 500K subset of BioTrove, providing natural-language captions that include visual information beyond species identity (the prompt for constructing these captions is in the supplemental materials).

\begin{table*}[t]
  \centering

  \hspace{-.26cm}\begin{subtable}[t]{0.75\textwidth}
    \centering
    \scalebox{\iliasScale}{%
        \begin{tabular}{@{}lllrrr@{}}
\toprule
\textbf{model}              & \textbf{backbone} & \textbf{resolution} & \textbf{I2I @ 100M}   & \textbf{T2I @ 5M}     & \textbf{T2I @ 100M} \\ \midrule
OAI CLIP           & ViT-B & 224  & 4.2          & 2.7          & 1.6        \\
OAI CLIP + {\n}    & ViT-B & 224  & 8.4 (+4.2)   & 11.5 (+8.8)  & 9.2 (+7.6) \\\midrule
SigLIP             & ViT-B & 512  & 16.6         & 14.6         & 11.1       \\
SigLIP + {\n}      & ViT-B & 512  & 24.6 (+8.0)  & 28.9 (+14.3) & 25.6 (+14.5) \\\midrule
SigLIP2            & ViT-B & 512  & 15.4         & 14.6         & 10.4       \\
SigLIP2 + {\n}     & ViT-B & 512  & 25.1 (+9.7)  & 30.3 (+15.7) & 27.1 (+16.7) \\\midrule
OpenCLIP           & ViT-L & 384  & 9.4          & 9.4          & 7.0        \\
OpenCLIP + {\n}    & ViT-L & 384  & 15.6 (+6.2)  & 20.9 (+11.5) & 18.7 (+11.7) \\\midrule
OAI CLIP           & ViT-L & 336  & 9.4          & 8.4          & 5.8        \\
OAI CLIP + {\n}    & ViT-L & 336  & 15.9 (+6.5)  & 20.6 (+12.2) & 18.9 (+13.1) \\\midrule
SigLIP             & ViT-L & 384  & 19.6         & 22.2         & 18.1       \\
SigLIP + {\n}      & ViT-L & 384  & 30.9 (+11.3) & 36.5 (+14.3) & 34.2 (+16.1) \\\midrule
SigLIP2            & ViT-L & 512  & 20.8         & 24.7         & 19.8       \\
SigLIP2 + {\n}     & ViT-L & 512  & 35.3 (+14.5) & 40.6 (+15.9) & 38.2 (+18.4) \\
\bottomrule
\end{tabular}
    }
    \caption{\label{tab:ilias_a}Comparison to baselines.}
  \end{subtable}\hfill
  \begin{subtable}[t]{0.265\textwidth}
    \centering
    \scalebox{\iliasScale}{%
        \begin{tabular}{@{}lr@{}}
        \toprule
        \textbf{model}              & \textbf{I2I @ 100M}   \\ \midrule
        OAI CLIP + TA             & 7.9           \\
        OAI CLIP + {\n}       & 21.0 (+13.1) \\ \midrule
        SigLIP + TA              & 23.0        \\
        SigLIP + {\n}        &  42.3 (+19.3) \\ \midrule
        SigLIP2 + TA            &  23.5        \\
        SigLIP2 + {\n}      &  43.2 (+19.7) \\ \midrule
        OpenCLIP + TA           & 13.7        \\
        OpenCLIP + {\n}         & 28.4 (+14.7) \\ \midrule
        OAI CLIP + TA            &15.2        \\
        OAI CLIP + {\n}        & 31.1 (+15.9) \\ \midrule
        SigLIP + TA             & 28.9        \\
        SigLIP + {\n}           & 45.8 (+16.9) \\ \midrule
        SigLIP2 + TA           & 31.3        \\
        SigLIP2 + {\n}          & 50.0 (+18.7) \\
        \bottomrule
        \end{tabular}
      }
    
    \caption{\label{tab:ilias_b}Comparison to static task adaptation (TA).}
  \end{subtable}
    \label{tab:ilias_results} 
  \caption{We show mAP@1k for image-to-image (I2I) and text-to-image (T2I) retrieval on ILIAS.
  }
  \label{tab:ilias_results}
\end{table*}

\paragraph{Metrics. }On ILIAS, we measure mean Average Precision @1k (mAP@1k) across both the full 100M distractor set (@100M) and the 5M mini distractor set (@5M). On INQUIRE, we measure mean Average Precision @50 (mAP@50), Normalized Discounted Cumulative Gain @50 (nDCG@50), and Mean Recall Rank (MRR). 

\paragraph{Baseline Models.} We use popular contrastively pretrained backbone models CLIP~\cite{radford2021learning}, SigLIP~\cite{Zhai_2023_ICCV_ogsiglip}, OpenCLIP~\cite{ilharco_gabriel_2021_5143773}, and SigLIP2~\cite{tschannen2025siglip2multilingualvisionlanguage} as backbone feature encoders. We replicate the baselines published with ILIAS for image-to-image re-ranking built on local feature descriptors from DINOv2~\cite{oquab2023dinov2}. These include query expansion-based methods like $\alpha$QE~\cite{Chum2007TotalRA_alphaqe}, local feature and geometric-based matching-based methods like Chamfer Similarity (CS) and Spatial Verification (SP)~\cite{wu2021densityaware_useschamfer}, and transformer-based methods like AMES~\cite{Suma_2024_ECCV_ames}. $\alpha$QE-$k$ refers to query expansion with $k$ nearest-neighbors. We also replicate baselines from INQUIRE using vision-language models (VLMs) as re-rankers, including open-source VLMs LLaVA~\cite{liu2023llava_neurips, liu2024llava15_cvpr, liu2025llava16_notes}, VILA~\cite{lin2024vila_cvpr}, 
PaliGemma~\cite{beyer2024paligemma}, InstructBLIP~\cite{dai2023instructblip_neurips}, and BLIP2~\cite{li2023blip2_icml}.

\section{Results}

\subsection{Embedding-based Retrieval}
\paragraph{ILIAS.} In Table~\ref{tab:ilias_a}, we report the retrieval performance using mAP@1k on the image-to-image and text-to-image tasks on ILIAS. For each baseline row, there is a corresponding row building {\n} on the same frozen baseline as a backbone feature encoder. Across all backbones, {\n} provides a strong performance boost from off-the-shelf encoders commonly used for retrieval.

\paragraph{Insufficiency of Static Task Adaptation. } 
A common method to improve the performance of large pretrained encoders is to adapt their features with a simple projection operation learned using a dataset that is relevant to the specific task~\cite{marchisio2023mini,rosenfeld2022ape}. While this allows models to be adapted for that general task, we show that learning a task adaptation performs significantly worse than {\n}'s query-specific adaptations. The authors of ILIAS show that a task adaptation trained on a sample of 1M images from the Universal Embeddings dataset~\citep{Ypsilantis_2023_ICCV} improves image-to-image instance retrieval in ILIAS 100M by between 3.7 and 10.5 mAP@1k. In Table~\ref{tab:ilias_b}, we compare this task adaptation approach (TA) with a version of {\n} that was also trained on a 1M image sample of the Universal Embeddings dataset. {\n} shows significant improvements, between 13.1 and 19.7 mAP@1k, on top of task adaptation improvements.

\begin{table}[t!]
\centering
\resizebox{.73\textwidth}{!}{
    \begin{tabular}{@{}lllrrr@{}}
        \toprule
        \textbf{model} & \textbf{backbone} & \textbf{resolution} & \textbf{mAP@50} & \textbf{nDCG@50} & \textbf{MRR} \\ \midrule
        OAI CLIP & ViT-B & 224 & 10.4 & 20.9 & 0.40 \\
        OAI CLIP + {\n} & ViT-B & 224 & 15.7 (+5.3) & 25.4 (+4.5) & 0.44 (+0.04) \\
        \midrule
        OAI CLIP & ViT-L & 336 & 23.4 & 37.7 & 0.59 \\
        OAI CLIP + {\n} & ViT-L & 336 & 32.5 (+9.1) & 43.6 (+5.9) & 0.64 (+0.05) \\
        \midrule
        SigLIP & ViT-L & 384 & 31.1 & 46.6 & 0.68 \\
        SigLIP + {\n} & ViT-L & 384 & 41.3 (+10.2) & 54.7 (+8.1) & 0.72 (+0.04) \\
        \midrule
        SigLIP & SoViT-400m & 384 & 34.2 & 49.1 & 0.69 \\
        SigLIP + {\n} & SoViT-400m & 384 & 45.4 (+11.2) & 56.8 (+7.7) & 0.74 (+0.05) \\
        \bottomrule
        \end{tabular}
    }

    \caption{Comparison of {\n} and baselines on INQUIRE}
    \label{tab:INQUIRE_MAIN}
\vspace{-.75em}
\end{table}

\begin{table}[t!]
\small \centering
\hspace{-.26cm}\resizebox{\textwidth}{!}{%
\begin{tabular}{@{}c@{\hspace{2em}}c@{}}
\begin{subtable}[t]{0.5\textwidth}
\centering
\begin{tabular}{lc}
\toprule
\textbf{Re-ranking Method} & \textbf{mAP@1k} \\
\midrule
Initial Ranking & 19.6 \\
\midrule
$\alpha$QE1 & 22.1 \\
$\alpha$QE2 & 20.4 \\
$\alpha$QE5 & 14.3 \\
\midrule
CS (Chamfer Similarity) & 22.9 \\
SP (Spatial Verification) & 21.8 \\
AMES & 26.4 \\
\midrule
{\n} & \textbf{29.1} \\
\bottomrule
\end{tabular}
\caption{\label{tab:ilias_reranking}ILIAS Top-1k Re-ranking}
\vspace{1.8cm}  %
\end{subtable}
&
\begin{subtable}[t]{0.6\textwidth}
\centering
\begin{tabular}{@{}lccc@{}}
\toprule
\textbf{Re-ranking Method} & \textbf{mAP@50} & \textbf{nDCG@50} & \textbf{MRR} \\
\midrule
Initial ranking & 33.3 & 48.8 & 0.69 \\
Best possible re-rank & 65.6 & 72.7 & 0.96 \\
\midrule
\multicolumn{4}{c}{\textit{Open‑source VLMs}} \\
\midrule
BLIP‑2 FLAN‑T5‑XXL & 31.2 & 46.5 & 0.58 \\
InstructBLIP‑T5‑XXL & 33.0 & 48.3 & 0.64 \\
PaliGemma‑3B‑mix‑448 & 35.6 & 50.6 & 0.68 \\
LLaVA‑1.5‑13B & 32.2 & 47.9 & 0.64 \\
LLaVA‑v1.6‑7B & 32.3 & 47.9 & 0.62 \\
LLaVA‑v1.6‑34B & 35.7 & 51.2 & 0.69 \\
VILA‑13B & 35.7 & 50.8 & 0.65 \\
VILA‑40B & 40.2 & 54.6 & 0.72 \\
\midrule
\multicolumn{4}{c}{\textit{Closed Source VLMs}} \\
\midrule
GPT‑4V & 36.5 & 51.9 & 0.72 \\
GPT‑4o & 43.7 &\textbf{57.9} & \textbf{0.78} \\
\midrule
SigLIP2 + {\n} & \textbf{45.6} & 55.1 & 0.76 \\
\bottomrule
\end{tabular}
\caption{\label{tab:inquire_reranking}INQUIRE Top-100 Re-ranking}
\end{subtable}
\end{tabular}%
}
\vspace{1pt}
\caption{Comparison of Re-ranking Methods on ILIAS and INQUIRE}
\label{tab:rerank}
\vspace{-0.75em}
\end{table}

\paragraph{INQUIRE.} Table~\ref{tab:INQUIRE_MAIN} shows retrieval performance across baseline model and backbone encoder sizes, along with corresponding {\n} models built upon their frozen features. In all cases, {\n} provides a significant performance improvement over general-purpose global features.

\subsection{Re-ranking}
In Table~\ref{tab:rerank}, we compare {\n} to other approaches for image-to-image re-ranking. We replicate the baselines used by ILIAS~\cite{kordopatiszilos2025iliasinstancelevelimageretrieval}, and pre-compute 100 local feature descriptors from DINOv2~\cite{oquab2023dinov2} for each gallery image, and 600 local feature descriptors for each query image. We evaluate these methods on the image-to-image re-ranking task over a set of top-1000 initial retrievals.

We demonstrate higher performance than baseline methods, most of which require computing and storing many local feature descriptors per query, while we only use global features that would typically already be stored in a database index.

We also evaluate {\n} on text-to-image re-ranking in INQUIRE against large VLMs. Table~\ref{tab:inquire_reranking} shows that {\n} outperforms all open-source VLMs on text-to-image re-ranking on this task while using precomputed global features and much lower computational overhead (explored further in Section~\ref{sec:computational_complexity}). {\n} is also competitive with closed-source VLMs GPT-4V and GPT-4o~\cite{openai2023gpt4v, openai2025gpt4o}.
\vspace{-.5em}\subsection{Ablation Studies}
We present an ablation study of {\n} in Table~\ref{tab:ABLATION}. We use the SigLIP2 backbone as the frozen feature extractor and evaluate both image-to-image and text-to-image tasks on ILIAS and INQUIRE. First, we compare the performance of the pretrained SigLIP2 model with models that are fine-tuned on the same datasets on which we train {\n}. This shows that fine-tuning alone, even on data from relevant domains, only provides a small improvement.

We then consider the performance of {\n} when we remove different components of the algorithm. First, we remove the iterative generation process and use a one-step generation process instead, resulting in a 6.8 decrease in mAP@1k on image-to-image retrieval @ 100M, and an 8.5 decrease in mAP@1k in text-to-image retrieval @ 5M on ILIAS, and a 6.9 decrease in mAP@50 on INQUIRE. Next, we remove the consideration of semi-positive samples used during training. This results in a degradation of 5.0 mAP@1k on both retrieval tasks on ILIAS and 4.8 mAP@50 on INQUIRE. Finally, we consider the case where noise is not added to the query representation during training to bridge the modality gap. This has the most significant impact on performance, with a drop of 15.3 image-to-image mAP@1k on ILIAS 100M, 8.2 text-to-image mAP@1k on ILIAS 5M, and 6.6 mAP@50 on INQUIRE. Notably, on the image-to-image task, {\n} without adding noise to the query representation during training does worse than the baseline SigLIP2 model, indicating that with only text-image data, this method could be prone to over-fitting without the additional regularization.

\begin{table*}[t!]
\small \centering
\resizebox{.85\textwidth}{!}{
\begin{tabular}{llllll}
\hline
\multicolumn{1}{c}{} & \multicolumn{2}{c}{\textbf{ILIAS}} & \multicolumn{3}{c}{\textbf{INQUIRE}} \\ \hline
\textbf{method} & \textbf{I2I @ 100M} & \textbf{T2I @ 5M} & \textbf{mAP@50} & \textbf{nDCG@50} & \textbf{MRR} \\ \midrule
SigLIP2 & 20.8 & 24.7 & 37.2 & 52.3 & 0.71 \\
Fine-tuned SigLIP2 & 21.1 & 25.1 & 38.9 & 53.1 & 0.72 \\\midrule
{\n} w/o Iterative Generation & 28.5 & 32.1 & 43.8 & 54.0 & 0.75 \\
{\n} w/o Semi-Positives & 30.3 & 35.6 & 45.9 & 53.8 & 0.75 \\
{\n} w/o Noise & 20.0 & 32.4 & 44.1 & 56.7 & 0.76 \\ \hline
{\n} & \textbf{35.3} & \textbf{40.6} & \textbf{50.7} & \textbf{58.3} & \textbf{0.79} \\ \hline
\end{tabular}
}

\caption{Ablation Studies on {\n}.}
\label{tab:ABLATION}
\vspace{-0.75em}
\end{table*}

\subsection{Embedding Visualizations}

\begin{figure}[t]
\small \centering
\resizebox{\textwidth}{!}{\begin{tabular}{cccc}
\textbf{Query} & \textbf{Original SigLIP2 Features} & \textbf{QuARI Adapted Features} & \\
\includegraphics[height=1.5in]{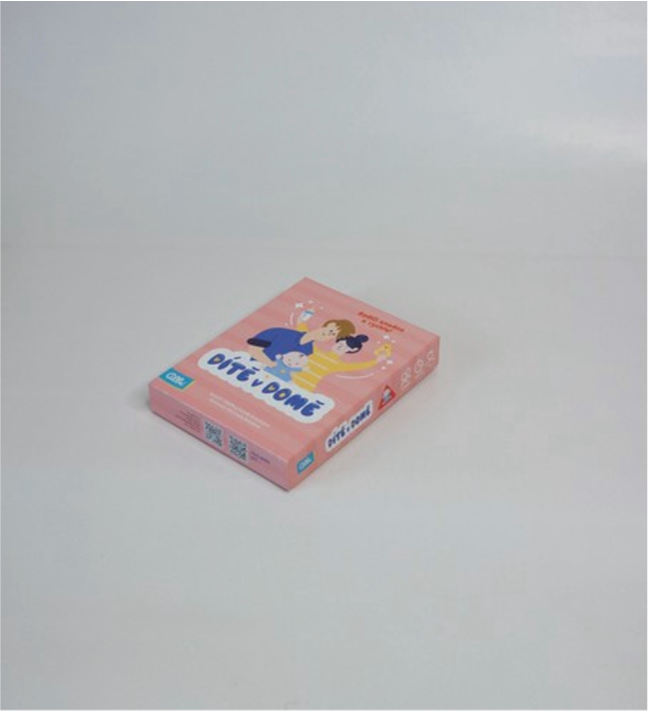} &
\includegraphics[height=1.5in]{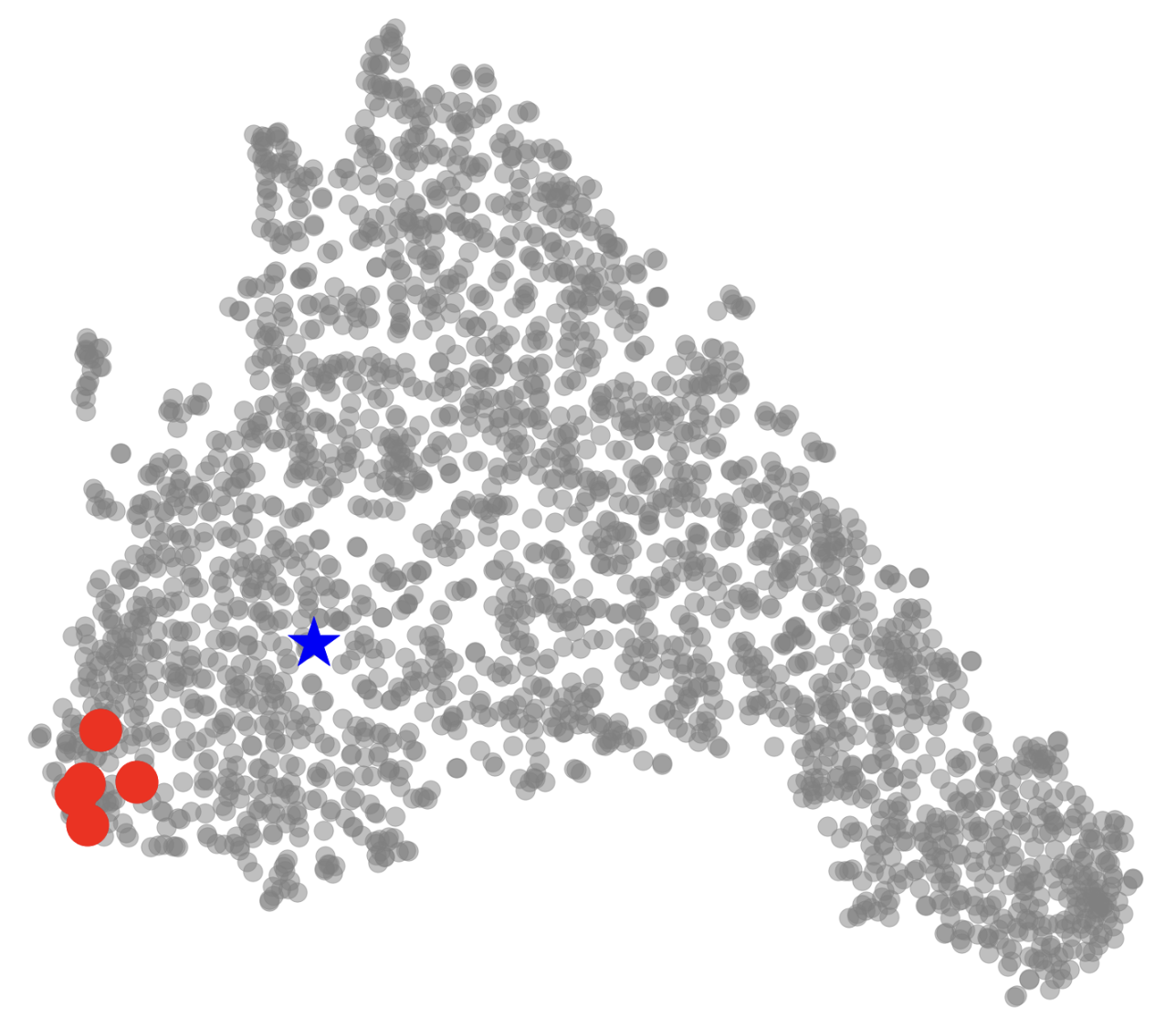} &
\includegraphics[height=1.5in]{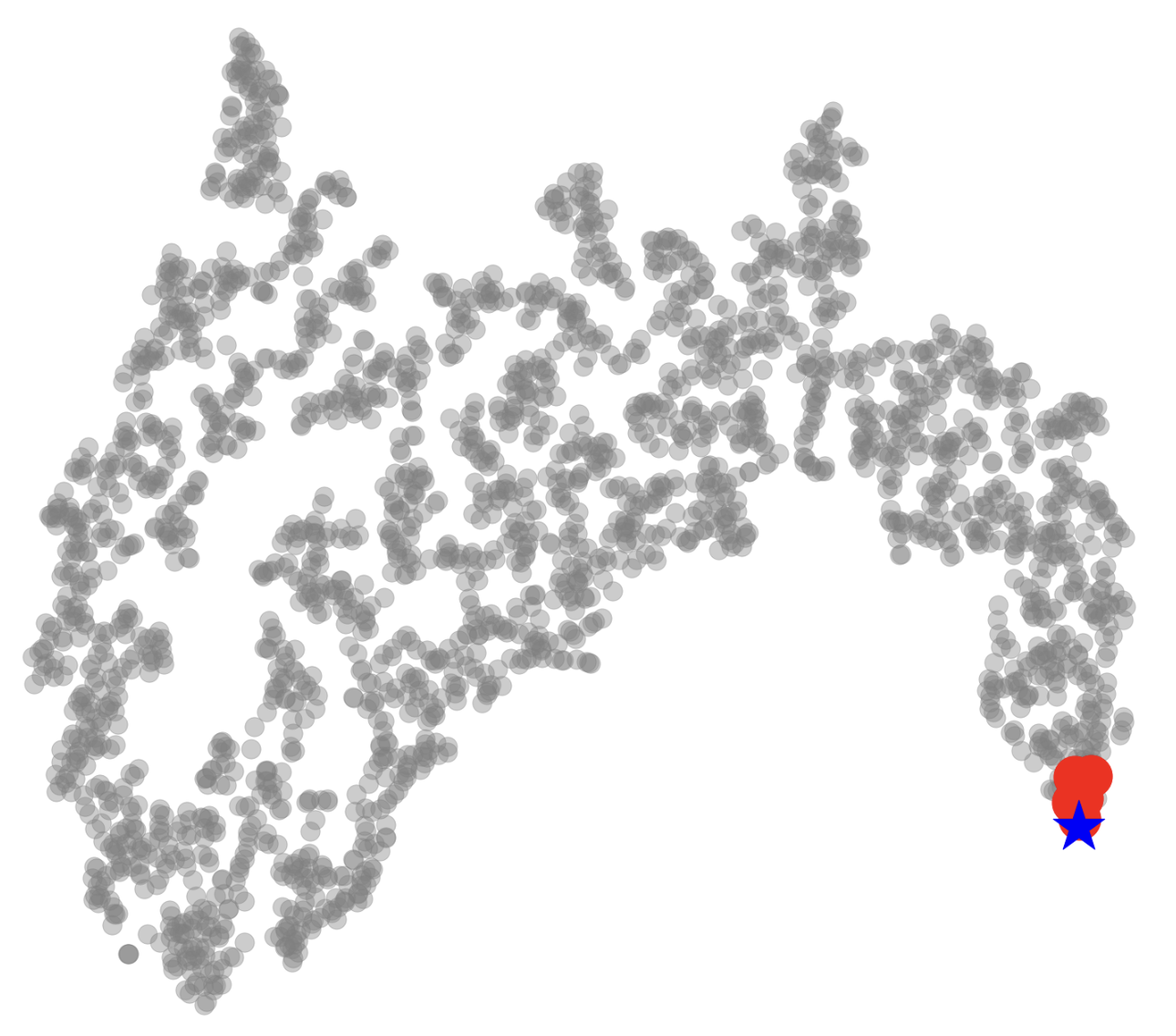} &
\hspace{-.45in}\raisebox{1.1in}{\includegraphics[height=0.32in]{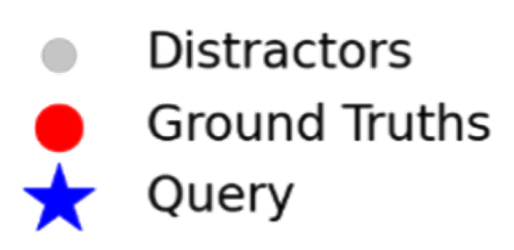}} \\ \hline

\centercell{\makecell[c]{``moray eel with open mouth\\ poking head out of burrows\\ or crevices''}} &
\raisebox{-.75in}{\includegraphics[height=1.5in]{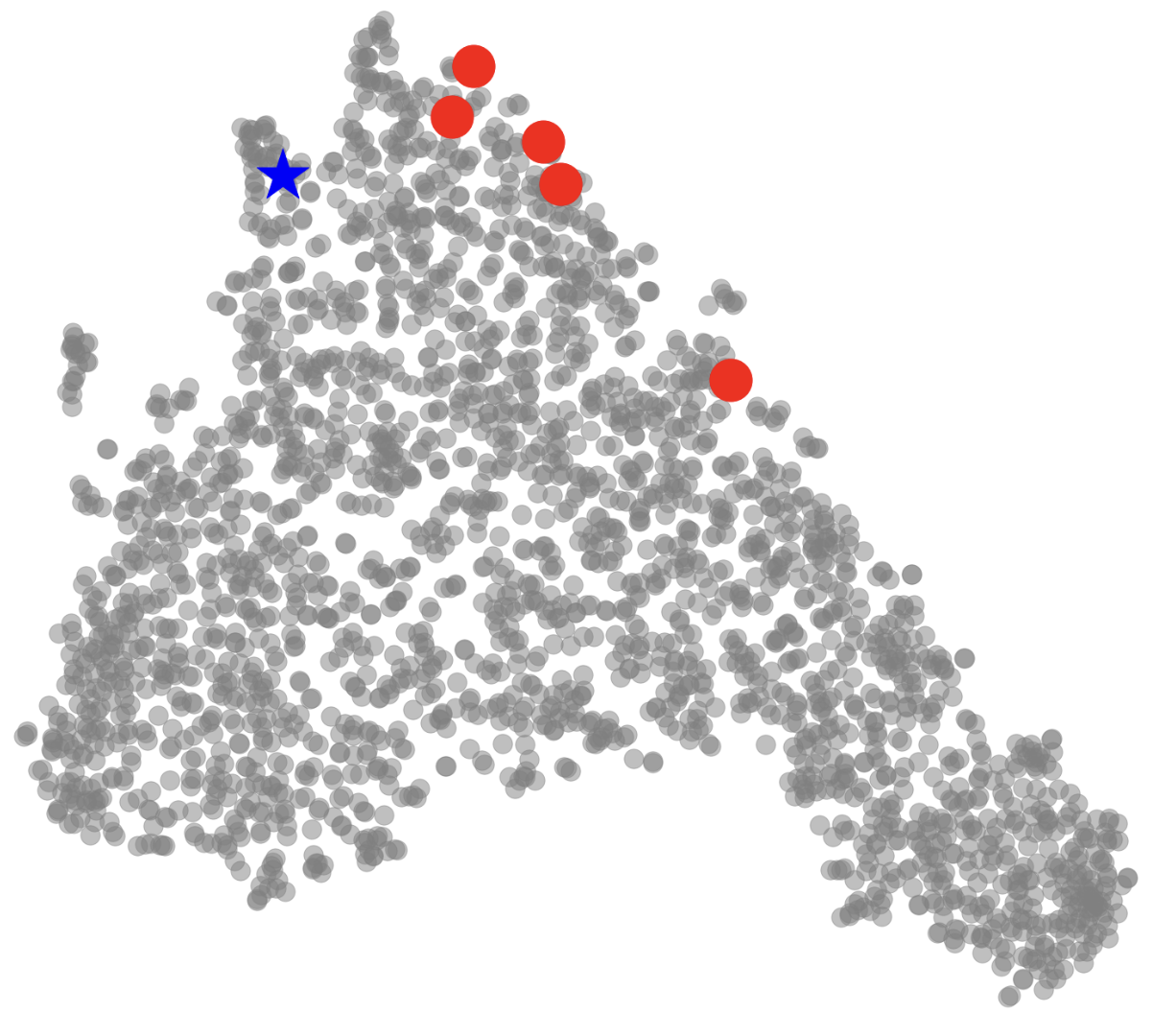}} &
\raisebox{-.75in}{\includegraphics[height=1.5in]{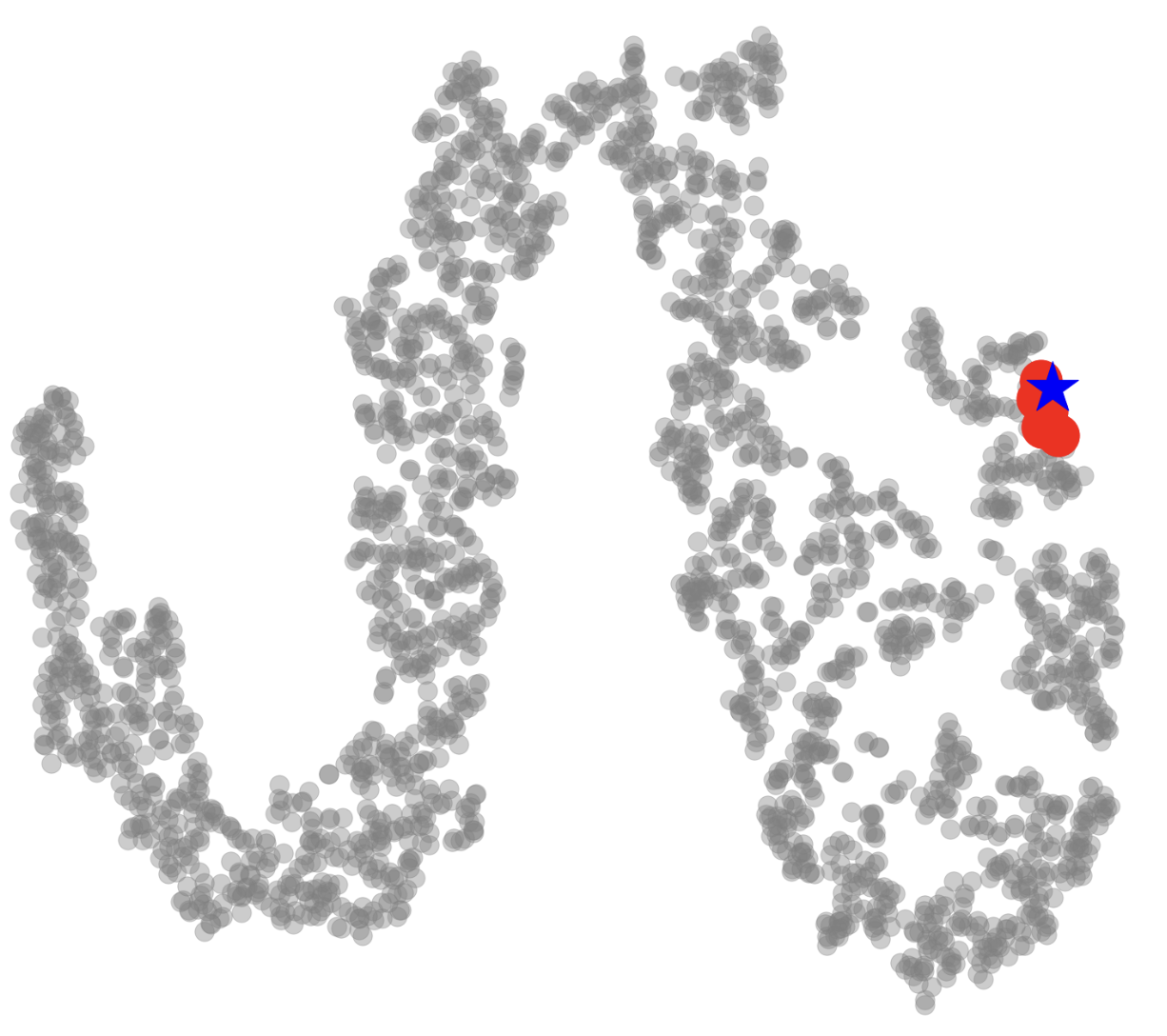}} &
\end{tabular}\vspace{1em}}
\caption{\label{fig:embeddings}t-SNE visualizations comparing original features and {\n} features.}\vspace{-0.75em}
\end{figure}

In Figure~\ref{fig:embeddings}, we explore feature transformations for two queries—an image query from ILIAS and a text query from INQUIRE. To visualize the original embedding space, the middle panel of each row shows the SigLIP2 embeddings for a collection composed of the two queries, their corresponding ground-truth images, and 5,000 distractor images (2,500 sampled at random from each of the ILIAS and INQUIRE datasets). In each row, the query embedding is highlighted in blue and the ground-truth responses in red. The right panel shows the t-SNE embedding of the same set after query-specific adaptation; here, the ground-truth responses are mapped much closer to their query embedding.

\subsection{Computational Efficiency}
\label{sec:computational_complexity}
\begin{figure}[t]
    \centering
    \begin{subfigure}[t]{0.48\textwidth}
        \centering
            
        \includegraphics[width=\linewidth,clip=true, trim={.55cm .55cm .55cm .55cm}]{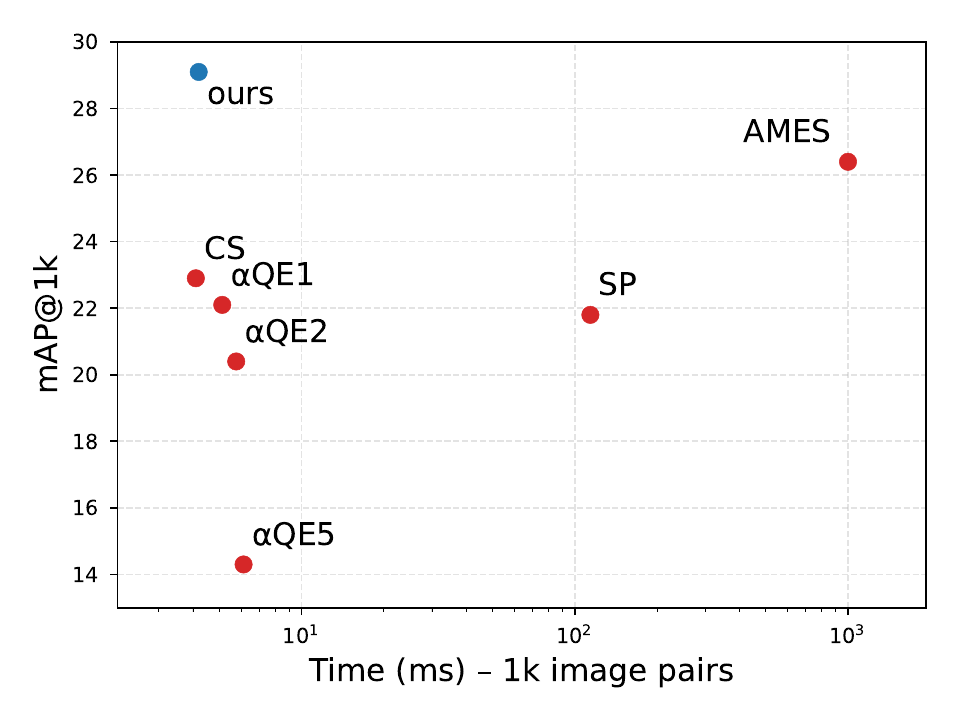}
    \end{subfigure}
    \hfill
    \begin{subfigure}[t]{0.48\textwidth}
        \centering
                \includegraphics[width=\linewidth,clip=true, trim={.55cm .55cm .55cm .55cm}]{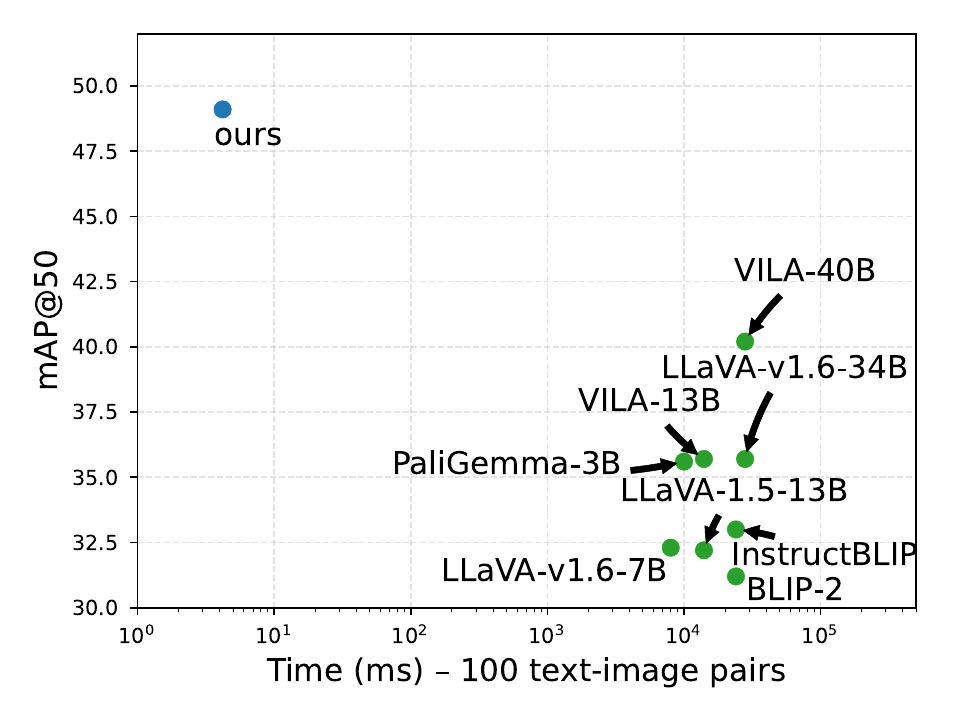}
    \end{subfigure}
    
    \caption{Comparison of re-ranking performance and inference cost for image-to-image retrieval on the ILIAS dataset (left) and text-to-image retrieval on the INQUIRE dataset (right). 
    }\vspace{-1em}
    
    \label{fig:rerank_summary}
\end{figure}

One of the primary strengths of {\n} is that it adapts features from a precomputed database of off-the-shelf features. Figure~\ref{fig:rerank_summary} shows the time to run a fixed-size query versus the performance of the method. {\n} not only achieves state-of-the-art performance, but is also very lightweight. On the ILIAS image-to-image re-ranking task, {\n} achieves around 3\% improvement over the highest accuracy re-ranking approach in over two orders of magnitude less time, and over 6\% better than approaches that have similar speeds \emph{without} the use of auxiliary local feature descriptors. On the INQUIRE text-to-image re-ranking task, {\n} is almost 10\% better than the best-performing vision-language model, and is orders of magnitude faster.

\section{Limitations} An inherent limitation of our method is that it applies a linear transformation to the retrieved results. While this design choice enables fast inference, it restricts the expressiveness of the adaptation. If the original representation space lacks relevant features for the retrieval, a linear transformation of it will be insufficient to improve results. Additionally, like most re-ranking approaches, our method is constrained by the set of top-$k$ results initially retrieved: it cannot recover relevant items that are excluded from this initial set. However, because {\n} is extremely efficient, this top-$k$ restriction is not as limiting in practice: we can afford to deploy {\n} over a very large initial set. Exploring non-linear transformation strategies to overcome representational limitations while maintaining computational tractability is a promising avenue for future research.

\section{Conclusions}
In this work, we introduced a query-specific retrieval framework, {\n}, that significantly outperforms strong baselines, including large vision-language models and models with learned domain-specific adaptations, on challenging retrieval benchmarks. By learning to predict transformations tailored to each query, our method significantly improves image-to-image and text-to-image retrieval performance without incurring substantial computational overhead. Unlike traditional re-ranking pipelines that rely on expensive local descriptors or multi-stage processing, our approach operates directly on global embeddings and scales efficiently for searching large databases of images. Our results demonstrate that retrieval performance can be meaningfully improved not by making the underlying encoders larger or more specialized, but by learning lightweight, query-conditioned adaptations of their features.

\clearpage
\appendix

\section{Implementation Details}
{\n's} transformer backbone is randomly initialized with 4 transformer layers. The query encoder and both the query and column decoders are two-layer MLPs with \texttt{GeLU} activation functions and layer normalization~\cite{ba2016layernormalization}. We train with a batch size of 320 and a contrastive temperature of 0.07. All backbone model embeddings are precomputed before training.

\section{Data Generation Prompt}
For all datasets other than BioTrove~\cite{yang2024biotrove}, we use the provided natural language annotations as the text label. However, BioTrove does not provide natural language annotations outside of taxonomic and common-name identities. Therefore, we provide the species annotation along with the image to \texttt{Qwen2.5-VL-7B-Instruct}~\cite{qwen2.5-VL} with the following instruction:

\emph{``For the image shown, write one plain, human-sounding sentence that someone might type into an image search system to find this exact picture of a \{species\_name\}. Mention the main objects, their key attributes, and any distinctive action or setting. Keep it brief and objective, avoiding flowery descriptors unless they are essential to identify the scene. Output only this sentence."}

We collect these annotations on 500K images sampled from BioTrove to augment our training dataset with natural language descriptions of biodiversity-domain imagery.

\section{Broader Impacts}
Improving retrieval systems to be both more accurate and more computationally efficient has broad positive implications, especially in domains where real-time or large-scale search is critical -- such as recognizing where victims of human trafficking are photographed~\citep{hotels50k}, monitoring biodiversity using camera trap images in ecological surveys~\citep{Beery_2018_ECCV}, or identifying the spread of disinformation through manipulated visual media~\citep{dang2024detecting}. QuARI enables high-quality retrieval even with limited resources, making advanced search capabilities more accessible in a wider range of applications. We do not foresee unique negative societal impacts associated with QuARI beyond those that already exist with general-purpose image retrieval systems. Nevertheless, the broader implications of visual search technologies—including potential misuse in surveillance or disinformation—remain important areas for ongoing community oversight and ethical consideration.

\bibliographystyle{plainnat}
\bibliography{main}

\end{document}